\documentclass[12pt,a4paper]{article}%
\usepackage{graphicx}
\usepackage{enumerate}
\usepackage{dcolumn}
\usepackage{multirow}
\usepackage{comment}
\usepackage{threeparttable}
\usepackage{booktabs,needspace,calc,makecell}
\usepackage[centertags,sumlimits, intlimits, namelimits,]{amsmath}
\usepackage{amsfonts}
\usepackage[english]{babel}
\usepackage{amssymb}
\usepackage{amsmath,amsthm}
\usepackage{natbib}
\usepackage{inputenc}
\usepackage{url}
\usepackage{longtable}
\usepackage[small]{titlesec}
\usepackage{chngcntr}

\usepackage{rotating}
\usepackage{mathrsfs}
\usepackage{yfonts}

\usepackage{longtable}
\usepackage{caption}
\usepackage{subcaption}
\usepackage{float}
\usepackage[usenames,dvipsnames]{xcolor}
\usepackage{color, colortbl}
\usepackage{varioref}

\usepackage{geometry}
\usepackage{gensymb}
\usepackage[page,header]{appendix}
\usepackage{titletoc}
\usepackage{hyperref}
\usepackage{lipsum}
\usepackage[export]{adjustbox}
\usepackage{soul}

\newtheorem{theorem}{Theorem}

\newtheorem{assumption}{Assumption}

\DeclareGraphicsExtensions{.eps,.pdf}

\title{Revealing Unobservables by Deep Learning: \\ Generative Element Extraction Networks (GEEN)}
\author{Yingyao Hu\thanks{Johns Hopkins University, Department of Economics, Johns Hopkins University, Wyman Park
Building 544E, 3400 N. Charles Street, Baltimore, MD 21218, (email: yhu@jhu.edu).}  \, Yang Liu\thanks{International Monetary Fund, 700 19th St NW, Washington DC 20431 (e-mail: yliu10@imf.org and jyao@imf.org).}, and  \, Jiaxiong Yao\footnotemark[2]}

\begin{document}

\maketitle

\begin{abstract}
Latent variable models are crucial in scientific research, where a key variable, such as effort, ability, and belief, is unobserved in the sample but needs to be identified. This paper proposes a novel method for estimating realizations of a latent variable $X^*$ in a random sample that contains its multiple measurements. With the key assumption that the measurements are independent conditional on $X^*$, we provide sufficient conditions under which realizations of $X^*$ in the sample are locally unique in a class of deviations, which allows us to identify realizations of $X^*$. To the best of our knowledge, this paper is the first to provide such identification in observation.  We then use the Kullback–Leibler distance between the two probability densities with and without the conditional independence as the loss function to train a Generative Element Extraction Networks (GEEN) that maps from the observed measurements to realizations of $X^*$ in the sample. The simulation results imply that this proposed estimator works quite well and the estimated values are highly correlated with realizations of $X^*$.  Our estimator can be applied to a large class of latent variable models and we expect it will change how people deal with latent variables. 
\end{abstract}

\section{Introduction}

Unobservables play a crucial role in scientific research because empirical researchers often encounter a discrepancy between what is described in a model and what is observed in the data. A typical example is the so-called hidden Markov models, where a series of latent variables are observed with errors in multiple periods under conditional independence assumptions. While there is a huge literature on the estimation of the model with latent variables (e.g.,\citet{aigner1984latent, bishop1998latent}), this paper focuses on the estimation of \emph{realizations} of the latent variable, which are not observed anywhere in the data. Suppose that the ideal data for the estimation of a model is an i.i.d. sample of $(X^1, X^2,...,X^k,X^*)$ \footnote{We use capital letters to stand for a random variable and lower case letters to stand for the realization of a random variable. For example, $f_{V}(v)$ stands for the probability density function of random variable $V$ with realization argument $v$, and $f_{V|U}(v|u)$ denote the conditional density of $V$ on $U$.} and that the researcher only observed $(X^1, X^2,...,X^k)$ in the sample. Generally, we consider $X^j, j=1...k,$ as multiple measurements of $X^*$. Under conditional independence assumptions, this paper provides a deep learning method to extract the common element $X^*$ from multiple observables $(X^1, X^2,...,X^k)$. We build a Generative Element Extraction Networks (GEEN) to reveal realizations or draws of $X^*$ to achieve a complete sample of $(X^1, X^2,...,X^k,X^*)$ in the sense that the generated draws are observationally equivalent to the true values in the sample.

This paper is different from the imputation method because the latent variable is not observed anywhere in the sample and needs to be identified. For example, true earnings of households are not observed anywhere in household survey data, but are of great interest to know. By contrast, imputation requires at least some observations of the underlying variable. 

Researchers have already applied deep generative models for data imputation. \citet{yoon2018gain} creatively use the Generative Adversarial Imputation Nets (GAIN) to provide an imputation method, in which missing values are estimated so that they are observationally equivalent to the observed values from the GAIN's perspective. \citet{yoon2018gain} creatively use the Generative Adversarial Imputation Nets (GAIN) to provide an imputation method, in which missing values are estimated so that they are observationally equivalent to the observed values from the GAIN's perspective.\citet{li2019misgan} also propose a GAN-based (\citet{goodfellow2014generative}) framework for learning from complex, high-dimensional incomplete data to impute missing data. \citet{mattei2019miwae} introduce the missing data importance-weighted autoencoder for training a deep latent variable model to handle missing-at-random data. \citet{nazabal2020handling} present a general framework for Variational Autoencoders (VAEs) (\citet{kingma2013auto}) that effectively incorporates incomplete data and heterogenous observations. \citet{muzellec2020missing} leverage optimal transport to define a loss function for missing value imputation. \citet{yoon2020gamin} propose a novel Generative Adversarial Multiple Imputation Network (GAMIN) for highly missing Data. In this literature, latent spaces are used to represent high-dimensional observations, but are not identifiable because their latent spaces may vary with parameter initialization. In addition, all the missing data models require true values to be partially observed. 

However, relatively little research has focused on estimating realizations of latent variables, which are unobserved, or completely missing. In the economics literature, Kalman filter and structural vector autoregressions have often been used to estimate the realizations of latent variables, such as potential output \citep{kuttner1994estimating}, natural rate of interest \citep{laubach2003measuring,holston2017measuring}, and natural rate of unemployment \citep{king2007search}, but the literature makes parametric assumptions about the dynamics of latent variables and thus belongs to the estimation of models with latent variables. 

In our setting, we argue that the conditional independence restrictions imply the local identification of the true values. That allows us to provide an estimator in the continuous case. Our method is nonparametric in the sense that we do not assume the distribution of the variables belong to a parametric family as in the widely-used VAEs \citep{kingma2013auto}, which use the so-called Evidence Lower Bound (ELBO) to provide a tractable unbiased Monte Carlo estimator. The VAEs focus on the estimation of a parametric model. In this paper, we focus on the estimation of the true values in each observation in the sample without imposing a parametric structure on the distributions. 

Our loss function is a distance between two nonparametric density functions with and without the conditional independence. Such a distance is based on a powerful nonparametric identification result in the measurement error literature \citet{hu2008instrumental}. (See \citet{hu2017econometrics} and \citet{schennach2020mismeasured} for a review.) It shows that the joint distribution of a latent variable and its measurements is uniquely determined by the joint distribution of the observed measurements under a key conditional independence assumption, together with other technical restrictions. To measure the distance between two density functions, we choose the Kullback–Leibler  divergence \citep{kullback1951information}, which plays a leading role in machine learning and neuroscience \citep{perez2008kullback}. A large literature has studied the estimation of the Kullback–Leibler  divergence \citep{darbellay1999estimation,moreno2003kullback, wang2005divergence,lee2006estimation, wang2006nearest, nguyen2010estimating,nowozin2016f,belghazi2018mutual}. We use a combination of a deep neural network and kernel density estimators to generate density functions with and without the conditional independence and then compute their divergence. 

In this paper, we make a further argument that the nonparametric identification of the latent variable distribution implies that the true values in the sample are locally separable in the continuous case. To the best of our knowledge, this paper is the first to provide such identification in observation. We expect such identification will change how researchers deal with latent variables and make our GEEN broadly applicable.

This paper is organized as follows. Section 2 provides the identification arguments. Section 3 describe the neural network and the algorithm. The Monte Carlo simulations are provided in Section 4. Section 5 summarizes the paper. Given the page limit, we put in the Online Appendix a high-level application in the estimation of fixed effects in panel data models.

\section{From identification in distribution to identification in observation}

We assume that a researcher observe the distribution of $\{X^1, X^2,...,X^k\}$ from a random sample. Putting the estimation of the population distribution $f_{X^1, X^2,...,X^k}$ from the random sample aside, we face a key identification challenge: How to determine the distribution $f_{X^1, X^2,...,X^k,X^*}$ from the observed distribution $f_{X^1, X^2,...,X^k}$? Here we use a general nonparametric identification result in the measurement error literature. We assume

\begin{assumption}
\label{assumption 3.0} 
There exists a random variable $X^*$ with support $\mathcal{X}^{ \ast }$ such that  
\begin{eqnarray}  \nonumber
&&f_{X^1, X^2,...,X^k,X^*} \\ \nonumber
&=& f_{X^1|X^*} f_{X^2|X^*} \times ... \times f_{X^k|X^*} f_{X^*} 
\end{eqnarray}\normalsize
\normalsize 
\end{assumption}

We may consider the observables $(X^1, X^2,...,X^k)$ as measurements of $X^*$. Here we use \citet{hu2008instrumental} to show the uniqueness of $f(X^1, X^2,...,X^k,X^*)$. We assume three of the $k$ measurements are informative enough for the results in \citet{hu2008instrumental}. We assume

\begin{assumption}
\label{assumption 3.1} The joint distribution of $(X^1, X^2,...,X^k,X^*)$ with $k \geq 3$ admits a bounded density with respect to the product measure of
some dominating measure defined on their supports. All marginal and conditional densities are also bounded.
\end{assumption}

Before introducing more assumptions, we define an integral operator corresponding to $f_{X^1\vert X^{ \ast }}$, which maps $f_{X^{ \ast }}$ over support $\mathcal{X}^{ \ast }$ to $f_{X^1}$ over support $\mathcal{X}^{ 1 }$. Suppose that we know both $f_{X^{ \ast }}$ and $f_{X^1}$ are bounded and integrable. We define $\mathcal{L}_{b n d}^{1} \left (\mathcal{X}^{ \ast }\right )$ as the set of bounded and integrable functions defined on $\mathcal{X}^{ \ast }$, i.e.,
\begin{eqnarray}\nonumber
&&\mathcal{L}_{b n d}^{1} \left (\mathcal{X}^{ \ast }\right ) \\ \nonumber
&=&\left \{g :\int _{\mathcal{X}^{ \ast }}\left \vert g (x^{ \ast })\right \vert   dx^{ \ast } <\infty \;\text{and}\;\sup_{x^{ \ast } \in \mathcal{X}^{ \ast }}\left \vert g (x^{ \ast })\right \vert  <\infty \right \}. \label{equ 060}
\end{eqnarray}\normalsize
\normalsize
The linear operator can be defined as
\begin{eqnarray}L_{X^1\vert X^{ \ast }} &  : & \mathcal{L}_{b n d}^{1} \left (\mathcal{X}^{ \ast }\right ) \rightarrow \mathcal{L}_{b n d}^{1} \left (\mathcal{X}^1\right ) \label{equ 070} \\
\left (L_{X^1\vert X^{ \ast }} h\right ) \left (x\right ) &  = & \int _{\mathcal{X}^{ \ast }}f_{X^1\vert X^{ \ast }} (x\vert x^{ \ast })h (x^{ \ast })  dx^{ \ast } . \nonumber \end{eqnarray}\normalsize  
\normalsize
In order to identify the unknown distributions, we need the observables to be informative so that the following assumptions hold.
\begin{assumption}
\label{assumption 3.2}The operators $L_{X^1\vert X^{ \ast }}$ and $L_{X^2\vert X^1}$ are injective.\protect\footnote{ $L_{X^2\vert X^1}$ is defined in the same way as $L_{X^1\vert X^{ \ast }}$ in equation (\ref{equ 070}).}
\end{assumption}

\begin{assumption}
\label{assumption 3.3}For all $\overline{x}^{ \ast } \neq \widetilde{x}^{ \ast }$ in $\mathcal{X}^{ \ast }$, the set $\left \{x^3 :f_{X^3\vert X^{ \ast }} \left (x^3\vert \overline{x}^{ \ast }\right ) \neq f_{X^3\vert X^{ \ast }} \left (x^3\vert \widetilde{x}^{ \ast }\right )\right \}$ has positive probability.
\end{assumption}

\begin{assumption}
\label{assumption 3.4}There exists a known functional $M$ such that $M\left [f_{X^1\vert X^{ \ast }} \left ( \cdot \vert x^{ \ast }\right )\right ] =x^{ \ast }$ for all $x^{ \ast } \in \mathcal{X}^{ \ast }$.
\end{assumption}
\noindent The functional $M$ may be the mean, mode, medium, or another quantile of the distribution $f_{X^1\vert X^{ \ast }} \left ( \cdot \vert x^{ \ast }\right ) $. The identification result may be summarized as follows:

\begin{theorem}\label{HuSc2008} 
\citet{hu2008instrumental} Under assumptions \ref{assumption 3.0}, \ref{assumption 3.1}, \ref{assumption 3.2}, \ref{assumption 3.3}, and \ref{assumption 3.4}, the joint distribution $f_{X^1, X^2,...,X^k}$ uniquely determines the joint distribution $f_{X^1, X^2,...,X^k,X^*}$, which satisfies 
\begin{eqnarray}\nonumber
&&f_{X^1, X^2,...,X^k,X^*} \\ \label{equ 200}
&=&f_{X^1|X^*} f_{X^2|X^*} \times ... \times f_{X^k|X^*} f_{X^*}. 
\end{eqnarray}\normalsize
\end{theorem}



In the remaining discussion, we still use the conditional independence in Equation (\ref{equ 200}) because we are interested in the common element $X^*$ across all the observables. This identification result implies that if we have qualified measurements $X^1$, $X^2$ and $X^3$, we are able to provide a consistent estimator of $f_{X^1, X^2,...,X^k,X^*}$ from a sample of $(X^1, X^2,...,X^k)$. 

\textbf{Identification in observation} Next, we argue that draws of $X^*$ are locally identified in the sense that there is no observationally equivalent uncorrelated deviation from these draws. 

Let  $X^*_i$ be a random draw of $X^*$ in observation $i$ and we define \textit{an uncorrelated deviation} from that draw as 
\begin{equation} \label{indep deviation}
X^*_i+\delta_i \quad \mbox{with} \quad E(X^*_i  \delta_i)=E( \delta_i)=0
\end{equation}\normalsize
\normalsize
where $(X^*_i,\delta_i)$ is a i.i.d. random draw from the joint distribution of $(X^*,\delta)$. 
Notice that if we replace $X^*_i$ with $X^*_i+\delta_i$ as the new common element, the variance of the common element becomes $var(X^*) + var(\delta)$. 
That means the variance of the uncorrelated deviation must be different from that of the original $X^*$, i.e., $var(X^*)$. The distribution of $X^*+\delta$ must be different from that of $X^*$. These two different distributions can not lead to the same observed distribution,  $f_{X^1, X^2,...,X^k}$, because Theorem \ref{HuSc2008} implies that  $f_{X^1, X^2,...,X^k}$ uniquely determines $f_{X^*}$, including its variance $var(X^*)$. In other words, $X^*+\delta$ and $X^*$ can not be observationally equivalent given the sample of $(X^1, X^2,...,X^k)$. Therefore, the draws of $X^*$ are locally identified in the following sense:

\begin{theorem} \label{Theorem id}
Suppose that the assumptions in Theorem \ref{HuSc2008} hold. Given an observed sample $\{X^1_i, X^2_i,...,X^k_i \}$, which is a subset of the infeasible full sample $\{X^1_i, X^2_i,...,X^k_i, X^*_i \}$, no uncorrelated deviation from latent draws $X^*_i$, defined in Equation (\ref{indep deviation}), is observationally equivalent to $X^*_i$.  
\end{theorem}

Notice that we only use the identified variance of the latent $X^*$ to make this argument. The results in Theorem \ref{HuSc2008} implies that all the moments of the latent $X^*$ are identified. Therefore, such a local identification result as in Theorem \ref{Theorem id} should hold for more general deviations than the uncorrelated deviations defined in Equation (\ref{indep deviation}).

Furthermore, we may look at this problem from a different angle. Suppose we insert generated draws $\hat{X}^*_i$ in the sample $\{X^1_i, X^2_i,...,X^k_i \}$ to obtain $\{X^1_i, X^2_i,...,X^k_i, \hat{X}^*_i \}$. And we also suppose that the conditional independence in Equation (\ref{equ 200}) holds with the generated draws, i.e.,
\begin{equation*} \nonumber
f_{X^1, X^2,...,X^k,\hat{X}^*} = f_{X^1|\hat{X}^*} f_{X^2|\hat{X}^*} \times ... \times f_{X^k|\hat{X}^*z} f_{\hat{X}^*}. 
\end{equation*}\normalsize

\noindent In this case, even if $\hat{X}^*_i$ is not equal to the true $X^*_i$ in the infeasible full sample $\{X^1_i, X^2_i,...,X^k_i, X^*_i \}$, our inserted $\hat{X}^*_i$ will be observationally equivalent to the true $X^*_i$ because Theorem \ref{HuSc2008} guarantees that the distributions  $f_{X^1|X^*}$,  $f_{X^2|X^*}$, and $ f_{X^3,...,X^k,X^*}$ are uniquely determined by the observed $f_{X^1, X^2,...,X^k}$. Even if $\hat{X}^*_i \neq X^*_i$, we can still correctly estimate $f_{X^1|X^*}$,  $f_{X^2|X^*}$, and $ f_{X^3,...,X^k,X^*}$ using sample $(X_i^1, X_i^2,...,X_i^k,\hat{X}^*_i)_{i=1,2,...,N}$ with inserted $\hat{X}^*_i$, instead of the true values $X^*_i$. 

In addition, Theorem \ref{Theorem id} implies that if we add a noise $\delta_i$ to the inserted $\hat{X}^*_i$, where $\delta_i$ is an uncorrelated deviation from $\hat{X}^*_i$, the conditional independence fails when $\hat{X}^*_i$ is replaced with $\hat{X}^*_i+\delta_i$. That means the inserted draws $\hat{X}^*_i$ are locally unique among uncorrelated deviations. 

The identification result in Theorem \ref{Theorem id} can be extended to the case where $\delta_i$ is uncorrelated with $X^*_i$ conditional on the observables $(X^1, X^2,...,X^k)$ because the conditional distribution $f_{X^*|X^1, X^2,...,X^k}$ is identified by Theorem \ref{HuSc2008}. We define \textit{a conditionally uncorrelated deviation} from $X^*_i$ as $X^*_i+\delta_i$ with 
\begin{equation} \label{cond indep deviation}
E(X^*_i \delta_i \, \vert \, X_i^1, X_i^2,...,X_i^k) = E( \delta_i \, \vert \, X_i^1, X_i^2,...,X_i^k) =0
\end{equation}\normalsize
\normalsize
where $(X^*_i ,\delta_i , X_i^1, X_i^2,...,X_i^k)$ is a i.i.d. random draw from their corresponding joint distribution. The variance, and therefore distribution, of $X^*_i+\delta_i $ conditional on $(X^1, X^2,...,X^k)$ is different from those of $f_{X^*|X^1, X^2,...,X^k}$. Theorem \ref{HuSc2008} implies that they must correspond to different $f_{X^1, X^2,...,X^k}$. Therefore, there is no observationally equivalent conditionally uncorrelated deviation from latent draws $X^*_i$. We summarize this extension as follows:
\begin{theorem} \label{Theorem id2}
Suppose that the assumptions in Theorem \ref{HuSc2008} hold. Given an observed sample $\{X^1_i, X^2_i,...,X^k_i \}$, which is a subset of the infeasible full sample $\{X^1_i, X^2_i,...,X^k_i, X^*_i \}$, no \textit{conditionally uncorrelated deviation} from latent draws $X^*_i$, defined in Equation (\ref{cond indep deviation}), is observationally equivalent to $X^*_i$. 
\end{theorem}

The identification results in Theorems \ref{Theorem id} and \ref{Theorem id2} are based on the identification of second moments of $f_{X^*}$ and $f_{X^*|X^1, X^2,...,X^k}$. The fact that the two distributions are  identified nonparametrically, instead of just for second moments, implies that identification in observation may hold for more general deviations than \textit{uncorrelated deviations} and \textit{conditionally uncorrelated deviations}. Furthermore, the continuity of $(X_i^1, X_i^2,...,X_i^k)$ also plays a role in the identification in observation. In the case where $(X_i^1, X_i^2,...,X_i^k)$ is discrete, a realization $(X_i^1, X_i^2,...,X_i^k)=(x^1, x^2,...,x^k)$ may have a positive probability. For all the observations with $(X_i^1, X_i^2,...,X_i^k)=(x^1, x^2,...,x^k)$, one may draw $X^*$ from $f_{X^*|X^1, X^2,...,X^k}$. Such a generated sample will have a sampling distribution converging to $f_{X^*, X^1, X^2,...,X^k}$ as the sample size goes to infinity. Therefore, the identification of $X^*_i$ is up to the permutation of the draws of $X^*$ in each observation with $(X_i^1, X_i^2,...,X_i^k)=(x^1, x^2,...,x^k)$.   In the case where $(X_i^1, X_i^2,...,X_i^k)$ is continuous, however, the event $(X_i^1, X_i^2,...,X_i^k)=(x^1, x^2,...,x^k)$ has a zero probability. Such a realization only appears once in the sample even as the sample size goes to infinity and this generated sample may not have a sampling distribution converging to $f_{X^*, X^1, X^2,...,X^k}$ as the sample size goes to infinity.

\textbf{Convergence argument and loss function} For an estimator of a finite-dimensional parameter, its consistency requires that the estimator should converge to its true value as the sample size goes to infinity. In this paper, we have a large number of unknown parameters, i.e., the realized value of $X^*$ in each observation. Here, we present sufficient conditions for the consistency of a large number of $\hat{X}^*_i $, rather than all $\hat{X}^*_i $.

Suppose our identification results suggest that our estimates $\hat{X}^*_i $ should have the same distribution (and variance) as $X^*_i$. Then the sample moments of $\hat{X}^*_i $ should converge to the true moments. In other words, we have
\begin{eqnarray} \label{2nd moment consistency}
\frac{1}{N} \sum_{i=1}^{N} (\hat{X}^*_i )^2 - \frac{1}{N} \sum_{i=1}^{N} (X^*_i )^2 = o_p(1)
\end{eqnarray}\normalsize
\normalsize
Such a condition implies the consistency of our estimator under following assumptions.

\begin{theorem} \label{consistency}
Suppose that the estimator $\hat{X}^*_i  = X^*_i + \delta_i$ for $i=1,2,...,N$ satisfies 
\begin{eqnarray} \label{condition -- deviation}
 \frac{1}{N} \sum_{i=1}^{N} X^*_i  \delta_i  = o_p(1).
 \end{eqnarray}
 \normalsize 
Then, the consistency of the sample moment  in Equation  \ref{2nd moment consistency}, 
implies that for any $\epsilon>0$, the sample proportion of large deviations goes to zero, i.e.,

$$P_N \left ( \left | \hat{X}^*_i - X^*_i \right |>\epsilon \right ) := \frac{1}{N} \sum_{i=1}^{N} I( |\delta_i| > \epsilon)= o_p(1)$$
\normalsize
\end{theorem}

\textbf{Proof:}
With  $\hat{X}^*_i  = X^*_i + \delta_i $, we have 

\begin{eqnarray*} 
&& \frac{1}{N} \sum_{i=1}^{N} (\hat{X}^*_i )^2 - \frac{1}{N} \sum_{i=1}^{N} (X^*_i )^2 \\
&=& 2\frac{1}{N} \sum_{i=1}^{N} X^*_i  \delta_i  + \frac{1}{N} \sum_{i=1}^{N} (  \delta_i )^2 \\
&=&  o_p(1) + \frac{1}{N} \sum_{i=1}^{N} (  \delta_i )^2 \\
&=& o_p(1)
\end{eqnarray*}
\normalsize
Therefore, 
 $$  \frac{1}{N} \sum_{i=1}^{N} (  \delta_i )^2 = o_p(1) $$ \normalsize
Furthermore, we have

\begin{eqnarray*} 
&& \frac{1}{N} \sum_{i=1}^{N} (  \delta_i )^2 > \epsilon^2  \frac{1}{N} \sum_{i=1}^{N}  I( |\delta_i| > \epsilon)  
\end{eqnarray*}
\normalsize

Therefore, for any $\epsilon>0$, we have 

$$\frac{1}{N} \sum_{i=1}^{N} I( |\delta_i| > \epsilon)= o_p(1)$$
\normalsize
$\qed$

This result implies that the estimator for each observation is consistent w.r.t the sampling distribution. Theorem \ref{consistency} does not guarantee the consistency of $\hat{X}^*_i $ in a given observation, but the probability of a randomly-drew estimator $\hat{X}^*_i$  being consistent should converge to one.

This result can be extended to more general deviations. For example, condition \ref{condition -- deviation} may be replaced with  

\begin{eqnarray} 
 \frac{1}{N} \sum_{i=1}^{N} X^*_i  \delta_i  = c \times  \frac{1}{N} \sum_{i=1}^{N} (  \delta_i )^2 + o_p(1).
 \end{eqnarray}\normalsize 
 \normalsize
where $c$ is a constant satisfying $c \neq -\frac{1}{2}$. That means the identification result remains in some cases where the deviations are correlated with the true values in the limit.


Finally, the discussion above implies that we can use a loss function measuring the distance between a general joint distribution
$p = f_{X^1, X^2,...,X^k,X^*}$ 
and a distribution satisfying conditional independence $p_{ci}= f_{X^1|X^*} f_{X^2|X^*}...f_{X^k|X^*}f_{X^*}
$
in order to search for latent draws $X^*_i$. A natural choice is the Kullback–Leibler divergence 

$$D_{KL} \left (p(x) || p_{ci}(x) \right ) = \int p(x) \ln  \left ( \frac{p(x)}{p_{ci}(x)} \right) dx.$$

\section{Generative Element Extraction Networks (GEEN)}
We build a Generative Element Extraction Network (GEEN), $G$, to generate the latent realizations of $X^*_i$ satisfying the conditional independence. Let $\vec{V}$ stand for the vector of draws of variable $V$ in the sample, i.e., 
$\vec{X}^*=(X^*_1,X^*_2,...,X^*_N)^T$ and 
$\vec{X}^j=(X^j_{1},X^j_{2},...,X^j_{N})^T$. 
We generate $\vec{\hat{X^*}}$ as follows: 

\begin{equation}
\vec{\hat{X^*}}=G(\vec{X}^1,\vec{X}^2,...,\vec{X}^k).
\end{equation}\normalsize
with $\vec{\hat{X^*}}=(\hat{X^*_{1}},\hat{X^*_{2}},...,\hat{X^*_{N}})^T$.
The deep neural network $G$ is trained to minimize the Kullback–Leibler divergence 
\begin{equation*}
\underset{G}{\min} \, D_{KL} \left (\hat{p} \, || \, \hat{p}_{ci} \right) \quad s.t.  \int x\hat{f}_{X^1|\hat{X}^*} (x|x^*)dx = x^*  
\end{equation*}

with
 $
\hat{p} = \hat{f}_{X^1, X^2,...,X^k,\hat{X}^*}
$ \normalsize
and 
$
\hat{p}_{ci}= \hat{f}_{X^1|\hat{X}^*} \hat{f}_{X^2|\hat{X}^*}...\hat{f}_{X^k|\hat{X}^*}\hat{f}_{\hat{X}^*}
$ \normalsize
where $\hat{f}$ are empirical distribution functions based on sample $(\vec{X}^1,\vec{X}^2,...,\vec{X}^k, \vec{\hat{X^*}})$. 
Furthermore, we have 

\begin{eqnarray} \nonumber
&& D_{KL} \left (\hat{p} \, || \, \hat{p}_{ci} \right)  \\
&&=  \frac{1}{N} \sum_{i=1}^N \ln  \left ( \hat{f}_{X^1, X^2,...,X^k,\hat{X}^*}(X^1_i, X^2_i,...,X^k_i,\hat{X}^*_i) \right)  \nonumber \\
&& - \frac{1}{N} \sum_{i=1}^N \ln  \left ( \hat{f}_{X^1|\hat{X}^*} (X^1_i|\hat{X}^*_i)\hat{f}_{X^2|\hat{X}^*}(X^2_i|\hat{X}^*_i) \right.  \nonumber\\
&&\left. ...\hat{f}_{X^k|\hat{X}^*}(X^k_i|\hat{X}^*_i)\hat{f}_{\hat{X}^*} (\hat{X}^*_i) \right) \label{equ kl_loss}
\end{eqnarray}
\normalsize
Notice that $G$ enters the loss function through  $\vec{\hat{X^*}}=(\hat{X^*_{1}},\hat{X^*_{2}},...,\hat{X^*_{N}})^T$ in density estimators. 
To be specific, we can have a kernel density estimator \footnote{One may choose other types of density estimators.}

\begin{equation*} 
 \hat{f}_{X^j|\hat{X}^*}(x|x^*)= \frac{\hat{f}_{X^j,\hat{X}^*}(x,x^*)}{\hat{f}_{\hat{X}^*}(x^*) } \end{equation*}\normalsize
\begin{equation*} 
 \hat{f}_{X^j,\hat{X}^*}(x,x^*)= \frac{1}{m } \sum_{i=1}^m \frac{K(( X^j_{i}-x)/h^j)}{h^j}\frac{K((\hat{X}^*_i-x^*)/h^*)}{h^*}
\end{equation*}
\begin{equation*} 
 \hat{f}_{\hat{X}^*}(x^*)= \frac{1}{m } \sum_{i=1}^m \frac{K(( \hat{X}^*_i-x^*)/h^*)}{h^*}
\end{equation*}

\begin{eqnarray*} \nonumber
&&\hat{f}_{X^1, X^2,...,X^k,\hat{X}^*}(x^1, x^2,...,x^k,x^*) \\ \nonumber
&=& \frac{1}{m } \sum_{i=1}^m \left ( \frac{K(( \hat{X}^*_i-x^*)/h^*)}{h^*}  \prod_{j=1}^k \frac{K(( X^j_{i}-x^j)/h^j)}{h^j} \right) \\ \label{equ 400d}
&& 
\end{eqnarray*}

where $h$ stands for bandwidths, $N$ the total sampled observations, $m$ the number of points in each observation and $k$ is the number of features. In the loss function defined in equation (\ref{equ kl_loss}), it requires more than one data point to estimate the kernel density function. As a result, unlike other use cases that one training point is enough to calculate its corresponding loss, we need to sample $m$ ($>1$) points as one observation to calculate its loss. For example, to build the training sample we sample with replacement $m$ points from the entire training data points and repeat $N$ times, and we end up with $N$ observations in our training sample. The same practice is followed to construct our validation and test samples. The kernel function $K(\cdot)$ can simply be the standard normal density function. For the bandwidth, we adopt the so-called Silverman's rule, i.e., $h^j = w \sigma^j N^{-1/5}$ where $ \sigma^j$ is the standard error of $X^j$, and $w$ is the window size that is determined by hyper parameters tuning. Similarly, we may take $h^* = w \sigma^* N^{-1/5}$, where $ \sigma^*$ is the standard error of $X^*$. 


In this paper, we experiment GEEN with multilayer perceptrons (MLPs), but this framework can be readily applied to other deep neural network architectures. In our simulations, we impose a convolution structure on $X^1$ so that the normalization condition can be simplified. The parameters of our deep neural network are estimated by minimizing the loss function:

\begin{equation*} 
\mbox{Loss} =
D_{KL} \left (\hat{p} \, || \, \hat{p}_{ci} \right)
+ \lambda \left| \frac{1}{N}\sum_{i=1}^N X^j_{1}-\frac{1}{N}\sum_{i=1}^N \hat{X}^*_i\right|^2\\ \label{equ dnn_loss}
\end{equation*}
\normalsize

Early stopping is applied when the loss does not improve for certain epochs in the validation sample. We do not use any information from true $X_i^*$ during training,  validation or hyper-parameter tuning. Instead we use the loss defined in the above equation for validation and true $X_i^*$ are only used for final testing.      

\section{Simulations}
This section presents the performance of our neural network through simulations. We generate the sample as follows:
\begin{eqnarray}
X^j_i = m^j(X^*_{i}) + \epsilon^j_i
\end{eqnarray}\normalsize
for $j=1,2,...,k$ and $i=1,2,...,N$.
Without loss of generality, we normalize
$
m^1(x)= x
$
and 
$ E[\epsilon^1 | X^*]=0. $
We pick distributions for $(\epsilon^1,...,\epsilon^k, X^*)$ and functions $(m^2,...,m^k)$ to generate a sample $(X^1,...,X^k, X^* )$. We then train $G$ using the observed sample $(\vec{X}^1,\vec{X}^2,...,\vec{X}^k)$ to generate $(\vec{X}^1,\vec{X}^2,...,\vec{X}^k, \hat{X^*})$. That is $\vec{\hat{X^*}} = G(\vec{X}^1,\vec{X}^2,...,\vec{X}^k).$
We check the performance of $G$ by calculating the correlation coefficient between $\vec{X^*}$ and $\vec{\hat{X^*}}$. 

We use a 6-layer with 10 hidden nodes fully connected neural network. The window size $w$ and normalization term $\lambda$ are tuned as hyper-parameters.  We use kernel functions to approximate their density functions. Theoretically if a distribution is normal, the best choice for $w$ used in the kernel function is 1, so to tune $w$ we choose the range from 0.5 to 2. To tune $\lambda$, we arbitrarily choose the range from 0.1 to 0.5. For every experiment, we run 25 times to evaluate the robustness of model performance on its initialization. 
For the baseline case, we use 

\begin{equation*}
\begin{split}
k &= 4 \\ 
m^1(x) &= x \\
m^2(x) &= \frac{1}{1+e^x} \\
m^3(x) &= x^2 \\
m^4(x) &= \ln (1 +e^x) \\
\end{split}
\qquad
\begin{split}
\epsilon^1 & =  N(0, 1) \\
\epsilon^2 & =  Beta(2, 2) - \frac{1}{2} \\
\epsilon^3 & =  Laplace(0, 1) \\
\epsilon^4 & =  Uniform(0, 1) - \frac{1}{2} \\
X^* &= N (0,4)
\end{split}
\end{equation*}

\begin{figure}[h!]
    \centering
    \includegraphics[width = 0.7\columnwidth]{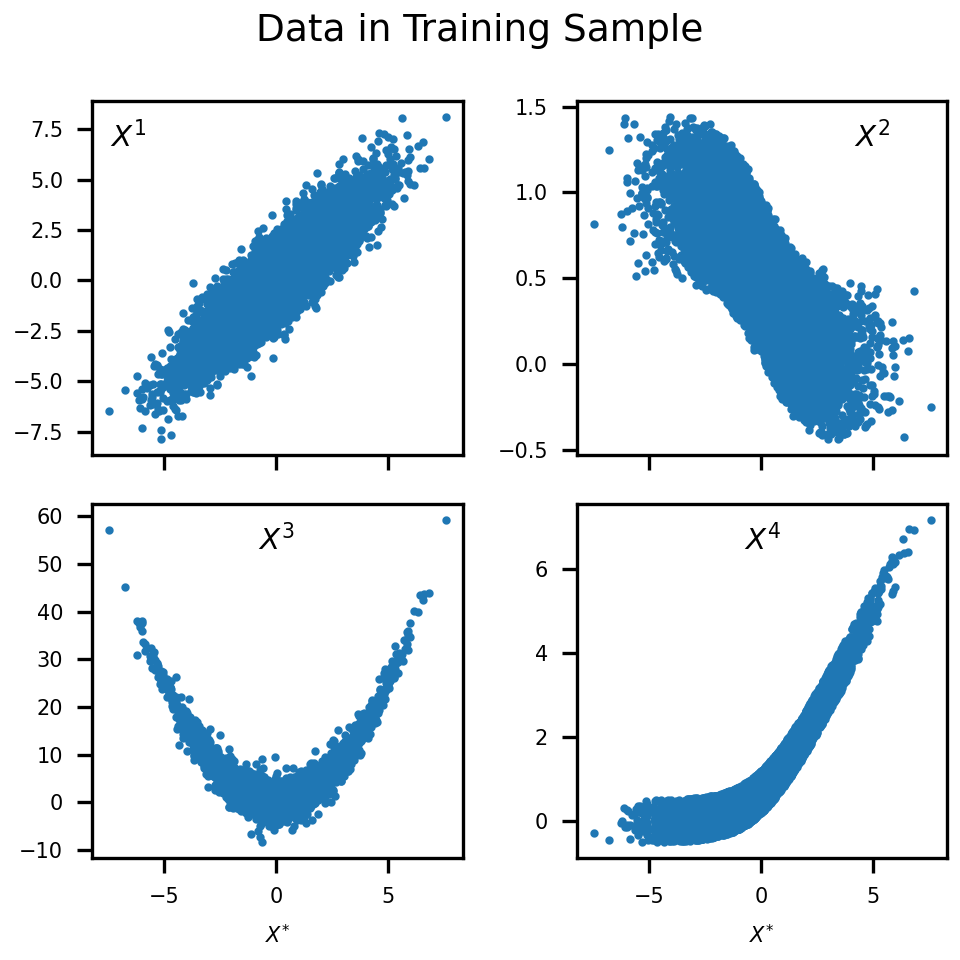}
    \caption{Baseline Training Sample}
    \label{fig:baseline_training_samples}
\end{figure}

We sample 8000 points as training points from the above distributions for $X^*$, $\epsilon^1$, $\epsilon^2$, $\epsilon^3$ and $\epsilon^4$. Then we sample another 1000 points for validation points and 1000 points for test points. We draw 500 points from the training points with replacement 8000 times to build our training set and 1000 times from the validation/test points to build our validation/test set. Figure \ref{fig:baseline_training_samples} shows the relationship between $X^1$, $X^2$, $X^3$, $X^4$ and $X^*$.  

\begin{figure}[h!]
    \centering
    \includegraphics[width = 0.7\columnwidth]{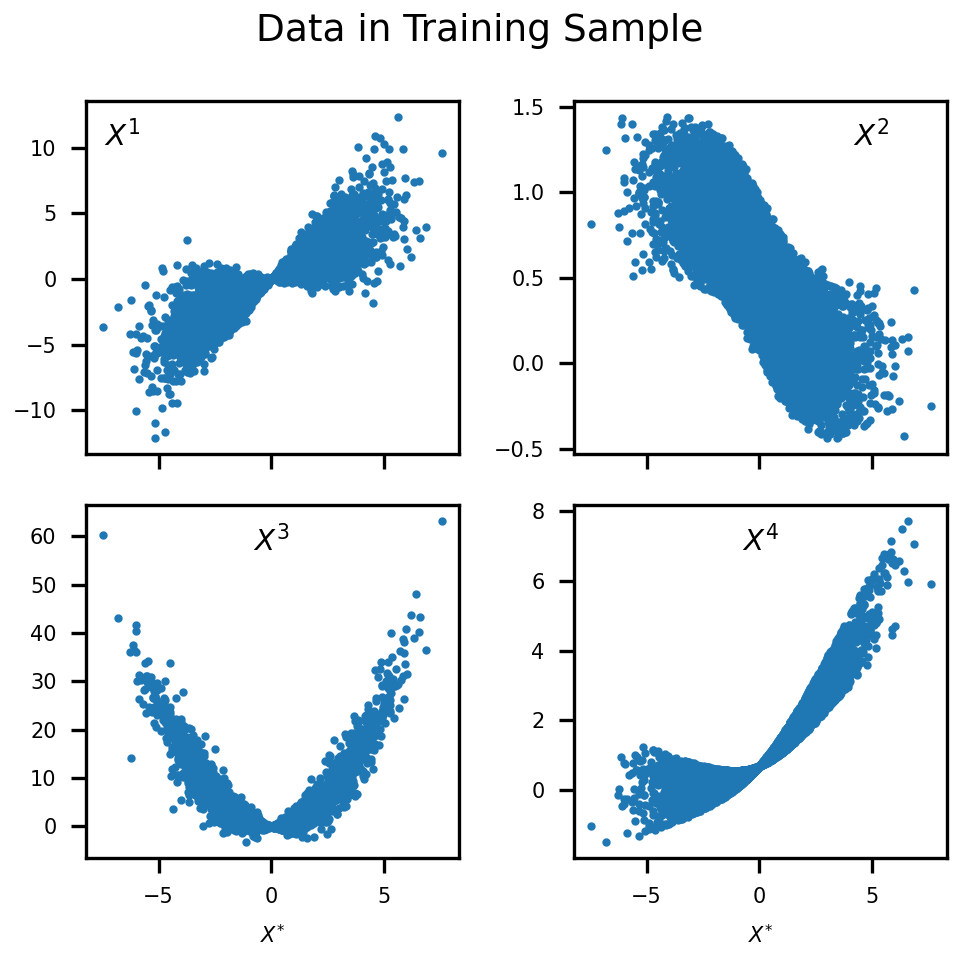}
    \caption{Linear Error Training Sample}
    \label{fig:linearError_training_samples}
\end{figure}

In the second experiment, we let the error terms correlate with $X^*$ while keeping the rest setup the same as the baseline. Figure \ref{fig:linearError_training_samples} shows the relationship between $X^1$, $X^2$, $X^3$, $X^4$ and $X^*$ in this one. Specifically, we use:

\begin{equation*}
\begin{split}
\epsilon^1 & =  N(0, \frac{1}{4}(x^*)^2) \\
\epsilon^4 & =  Uniform(0, \frac{1}{2}|x^*|) - \frac{1}{4}|x^*|
\end{split}
\quad
\begin{split}
\epsilon^3 & =  Laplace(0, \frac{1}{2}|x^*|) \\
&
\end{split}
\end{equation*}

\begin{figure}[h!]
    \centering
    \includegraphics[width = 0.7\columnwidth]{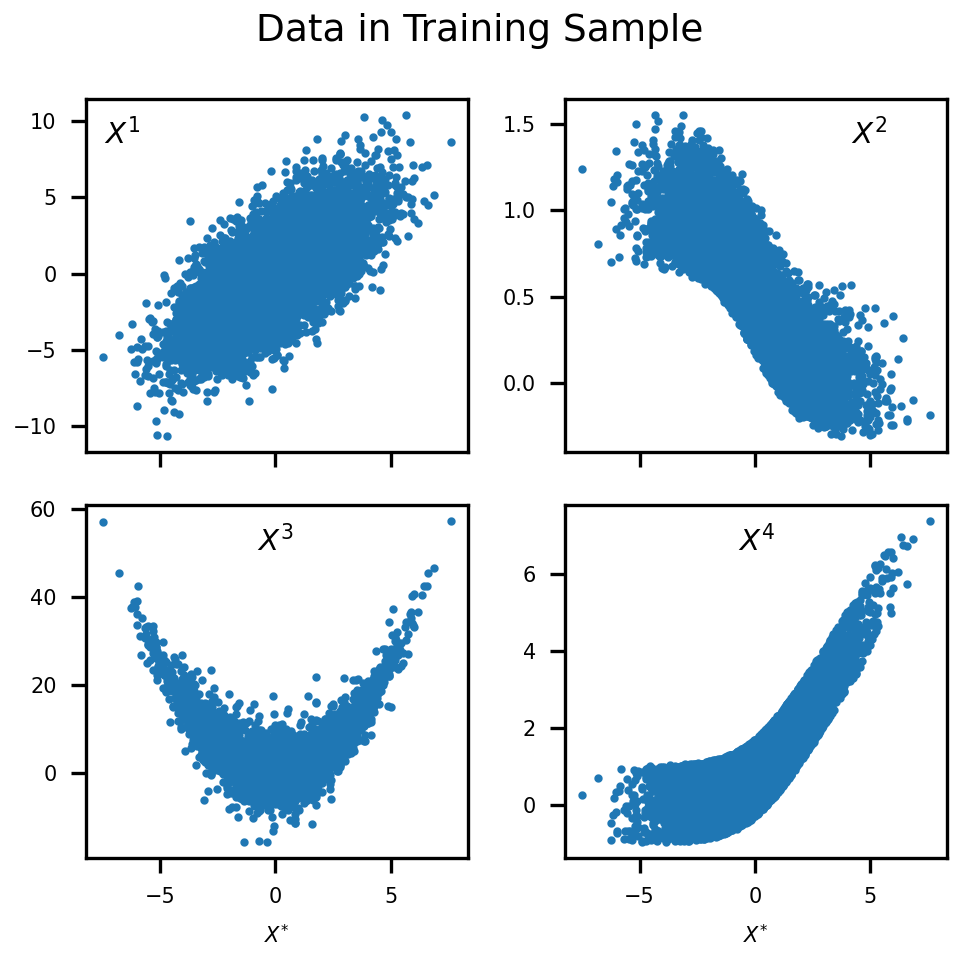}
    \caption{Double Error Training Sample}
    \label{fig:largeError_training_samples}
\end{figure}

In the third experiment, we double the variance of the error terms while keeping the rest setup the same as the baseline. Figure \ref{fig:largeError_training_samples} shows the relationship between $X^1$, $X^2$, $X^3$, $X^4$ and $X^*$ in the third experiment.

\begin{equation*}
\begin{split}
\epsilon^1 & =  N(0, 4) \\
\epsilon^2 & =  Beta(2, 4) - \frac{1}{3}
\end{split}
\qquad\qquad
\begin{split}
\epsilon^3 & =  Laplace(0, 2) \\
\epsilon^4 & =  Uniform(0, 2) - 1
\end{split}
\end{equation*}

\begin{figure}[!h]
    \centering
    \includegraphics[width = 0.7\columnwidth]{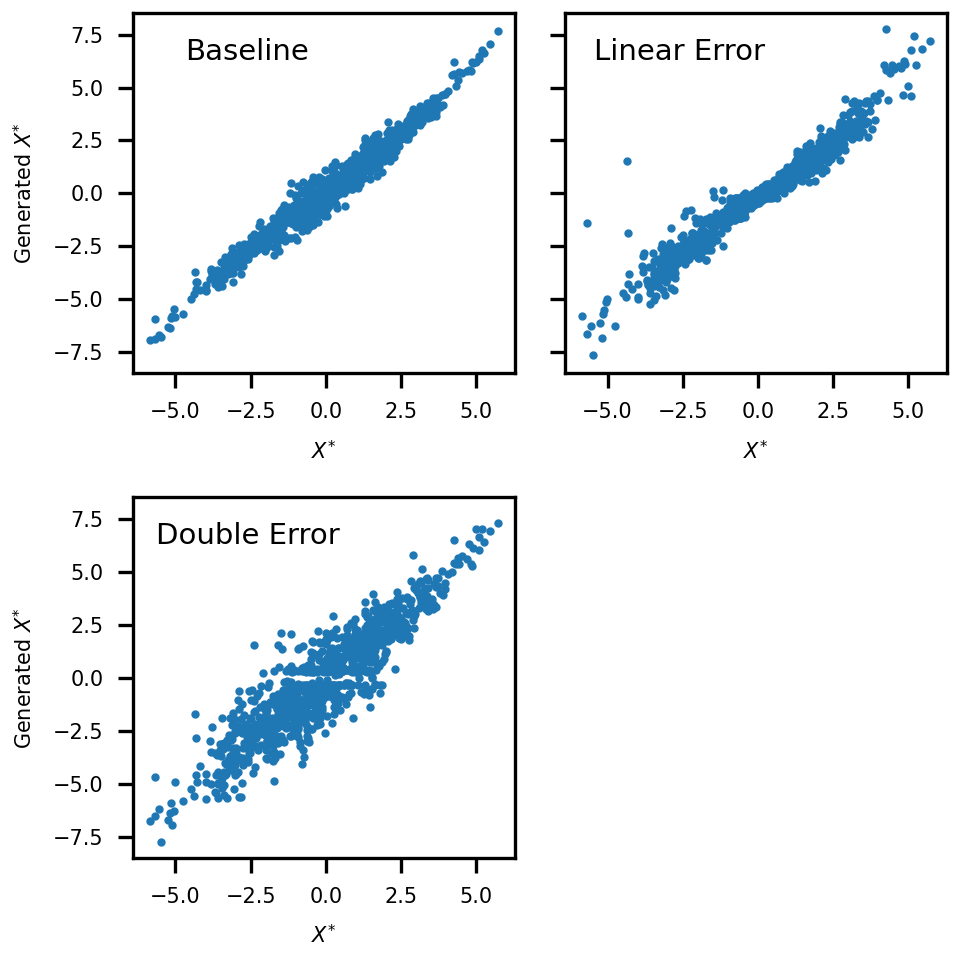}
    \caption{Results in the First Three Experiments}
    \label{fig:results}
\end{figure}

Table \ref{table:simmulation_summary} demonstrates the min, median and max correlations of $\vec{X^*}$ and $\vec{\hat{X^*}}$ in the test sample for the three experiments after running each one 25 times. Figure \ref{fig:results} shows their best runs respectively. It is clear that GEEN is robust with initialization with very tight distributions of the correlations of $\vec{X^*}$ and $\vec{\hat{X^*}}$ and improves significantly if we simply use $X^1$ to directly measure $X^*$. 

\begin{table}[h!]
\centering
\resizebox{0.7\textwidth}{!} {
{
\begin{threeparttable}
\begin{tabular}{l*{4}{c}}
 \toprule
 Simulation Name &\multicolumn{3}{c}{corr($\vec{X^*}$, $\vec{\hat{X^*}}$)}
 & corr($\vec{X^*}$, $\vec{X^1}$)\\
  & min & median & max  & \\
 \cmidrule{2-4}
 Baseline  & 0.97  & 0.98 & 0.98 & 0.89  \\

Linear Error  & 0.94  & 0.96 & 0.97 & 0.89  \\

Double Error  & 0.88  & 0.89 & 0.91 & 0.70  \\
\bottomrule
\end{tabular}
\end{threeparttable}
}
}
\caption[Summary of Simulation Results]{Summary of Simulation Results\label{table:simmulation_summary}}
\end{table}


\begin{figure}[!h]
    \centering
    \includegraphics[width =0.7\columnwidth]{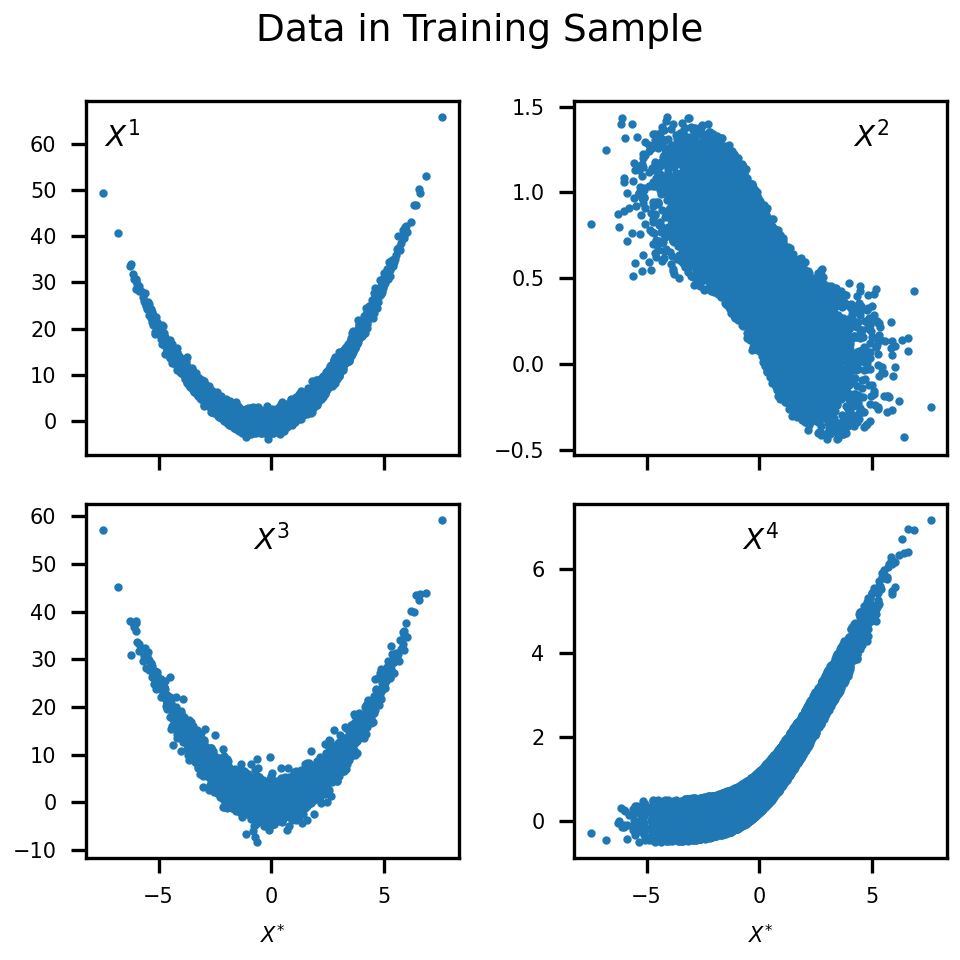}
    \caption{No Normalization Training Sample}
    \label{fig:woNormal_training_samples}
\end{figure}

In the forth experiment, we loose the normalization condition while keeping the rest setup the same as the baseline. Figure \ref{fig:woNormal_training_samples} shows the relationship between $X^1$, $X^2$, $X^3$, $X^4$ and $X^*$ in this experiment.
\begin{eqnarray*}
m^1(x) &=& x^2 + x 
\end{eqnarray*}\normalsize 

\begin{figure}[!h]
    \centering
    \includegraphics[width =0.7\columnwidth]{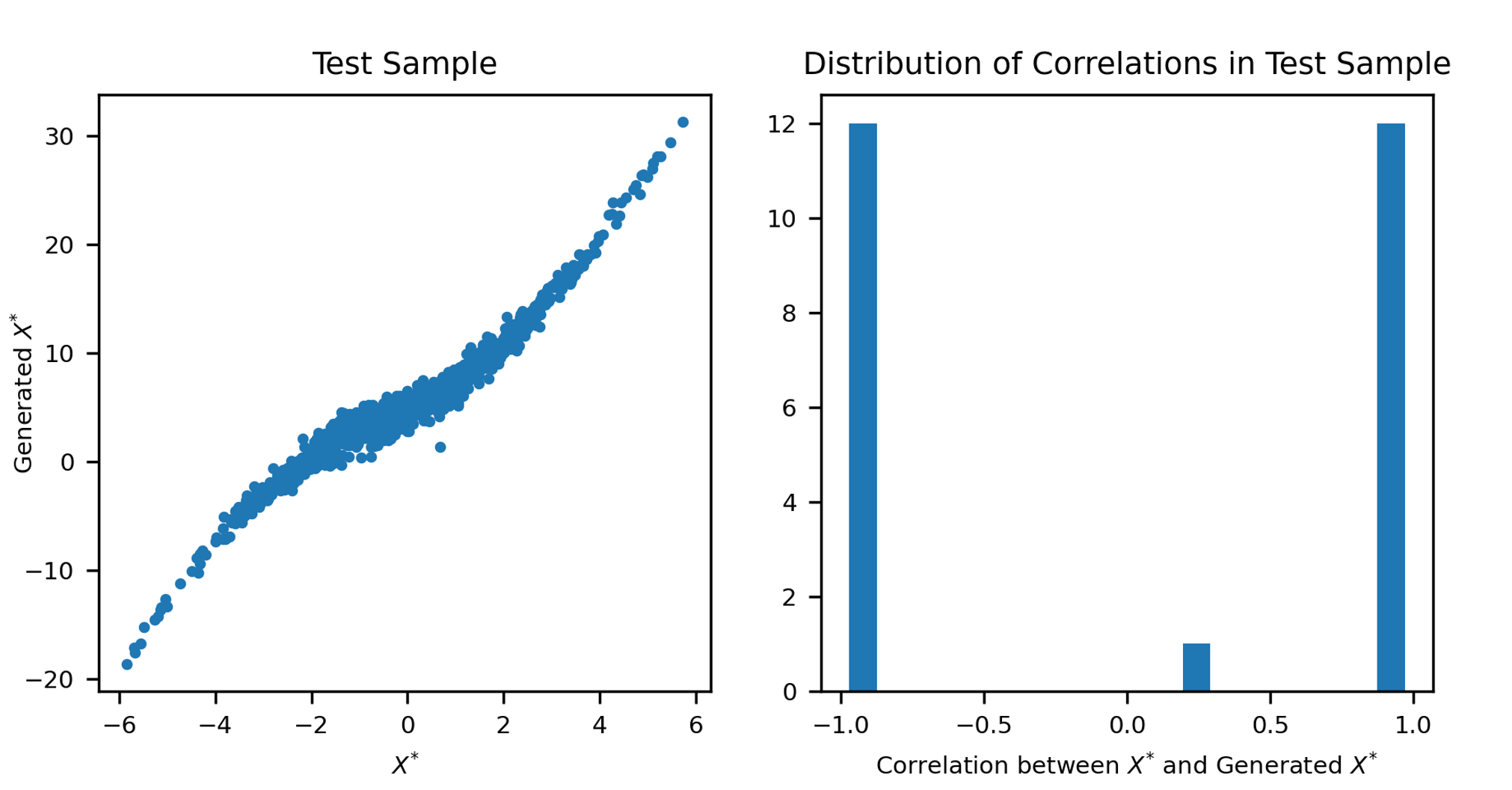}
    \caption{Results in No Normalization Experiment}
    \label{fig:woNormal_result}
\end{figure}

With this setup, generated $\hat{X^*}$ is not anchored, and as shown in the left hand side of Figure \ref{fig:woNormal_result} its values deviate significantly from $X^*$. However, the KL loss helps keep the similarity of the two distributions of generated $\hat{X^*}$ and $X^*$. As shown in the right hand side of Figure \ref{fig:woNormal_result}, with 25 runs of this experiment most of the absolute values of the correlation between $\vec{X^*}$ and $\vec{\hat{X^*}}$ are around 0.9. This suggests that even without normalization our framework can still help provide an estimation of the direction. 

\pagebreak

\section{Summary}

This paper proposes a Generative Element Extraction Networks (GEEN) to reveal unobserved variables in a random sample, which contains multiple measurements of a latent variable of interest. We present the sufficient conditions, under which the joint distribution of a latent variable and its measurements can be uniquely determined. We then argue that the true values of the latent variable in the sample are locally unique in a class of deviations, which allows us to estimate the true values. To the best of our knowledge, this paper is the first to provide such identification in observation.  This approach is based on the key assumption that the measurements are independent conditional on the latent variable. We then propose an algorithm to minimize the probability densities with and without the conditional independence. We use the Kullback–Leibler distance between the two probability densities as the loss function to train the deep neutral network, which maps from the observed measurements to the true values of the latent variable in the sample. The simulation results imply that this proposed estimator works quite well and the estimated values are highly correlated with the true values with a correlation coefficient usually higher than 90\%. We then apply this estimator to a panel data model to reveal the fixed effects (see our online appendix), which we believe is highly applicable to studies using longitudinal data. We expect this algorithm will change how researchers deal with latent variables in empirical research.

\pagebreak

\bibliographystyle{aaai23}
\bibliography{aaai23}

\begin{thebibliography}{27}
\providecommand{\natexlab}[1]{#1}

\bibitem[{Aigner et~al.(1984)Aigner, Hsiao, Kapteyn, and
  Wansbeek}]{aigner1984latent}
Aigner, D.~J.; Hsiao, C.; Kapteyn, A.; and Wansbeek, T. 1984.
\newblock Latent variable models in econometrics.
\newblock \emph{Handbook of econometrics}, 2: 1321--1393.

\bibitem[{Belghazi et~al.(2018)Belghazi, Baratin, Rajeshwar, Ozair, Bengio,
  Courville, and Hjelm}]{belghazi2018mutual}
Belghazi, M.~I.; Baratin, A.; Rajeshwar, S.; Ozair, S.; Bengio, Y.; Courville,
  A.; and Hjelm, D. 2018.
\newblock Mutual information neural estimation.
\newblock In \emph{International conference on machine learning}, 531--540.
  PMLR.

\bibitem[{Bishop(1998)}]{bishop1998latent}
Bishop, C.~M. 1998.
\newblock Latent variable models.
\newblock In \emph{Learning in graphical models}, 371--403. Springer.

\bibitem[{Darbellay and Vajda(1999)}]{darbellay1999estimation}
Darbellay, G.~A.; and Vajda, I. 1999.
\newblock Estimation of the information by an adaptive partitioning of the
  observation space.
\newblock \emph{IEEE Transactions on Information Theory}, 45(4): 1315--1321.

\bibitem[{Goodfellow et~al.(2014)Goodfellow, Pouget-Abadie, Mirza, Xu,
  Warde-Farley, Ozair, Courville, and Bengio}]{goodfellow2014generative}
Goodfellow, I.; Pouget-Abadie, J.; Mirza, M.; Xu, B.; Warde-Farley, D.; Ozair,
  S.; Courville, A.; and Bengio, Y. 2014.
\newblock Generative adversarial nets.
\newblock \emph{Advances in neural information processing systems}, 27.

\bibitem[{Holston, Laubach, and Williams(2017)}]{holston2017measuring}
Holston, K.; Laubach, T.; and Williams, J.~C. 2017.
\newblock Measuring the natural rate of interest: International trends and
  determinants.
\newblock \emph{Journal of International Economics}, 108: S59--S75.

\bibitem[{Hu(2017)}]{hu2017econometrics}
Hu, Y. 2017.
\newblock The econometrics of unobservables: Applications of measurement error
  models in empirical industrial organization and labor economics.
\newblock \emph{Journal of econometrics}, 200(2): 154--168.

\bibitem[{Hu and Schennach(2008)}]{hu2008instrumental}
Hu, Y.; and Schennach, S.~M. 2008.
\newblock Instrumental variable treatment of nonclassical measurement error
  models.
\newblock \emph{Econometrica}, 76(1): 195--216.

\bibitem[{King and Morley(2007)}]{king2007search}
King, T.~B.; and Morley, J. 2007.
\newblock In search of the natural rate of unemployment.
\newblock \emph{Journal of Monetary Economics}, 54(2): 550--564.

\bibitem[{Kingma and Welling(2013)}]{kingma2013auto}
Kingma, D.~P.; and Welling, M. 2013.
\newblock Auto-encoding variational bayes.
\newblock \emph{arXiv preprint arXiv:1312.6114}.

\bibitem[{Kullback and Leibler(1951)}]{kullback1951information}
Kullback, S.; and Leibler, R.~A. 1951.
\newblock On information and sufficiency.
\newblock \emph{The annals of mathematical statistics}, 22(1): 79--86.

\bibitem[{Kuttner(1994)}]{kuttner1994estimating}
Kuttner, K.~N. 1994.
\newblock Estimating potential output as a latent variable.
\newblock \emph{Journal of business \& economic statistics}, 12(3): 361--368.

\bibitem[{Laubach and Williams(2003)}]{laubach2003measuring}
Laubach, T.; and Williams, J.~C. 2003.
\newblock Measuring the natural rate of interest.
\newblock \emph{Review of Economics and Statistics}, 85(4): 1063--1070.

\bibitem[{Lee and Park(2006)}]{lee2006estimation}
Lee, Y.~K.; and Park, B.~U. 2006.
\newblock Estimation of Kullback--Leibler divergence by local likelihood.
\newblock \emph{Annals of the Institute of Statistical Mathematics}, 58(2):
  327--340.

\bibitem[{Li, Jiang, and Marlin(2019)}]{li2019misgan}
Li, S. C.-X.; Jiang, B.; and Marlin, B. 2019.
\newblock Misgan: Learning from incomplete data with generative adversarial
  networks.
\newblock \emph{arXiv preprint arXiv:1902.09599}.

\bibitem[{Mattei and Frellsen(2019)}]{mattei2019miwae}
Mattei, P.-A.; and Frellsen, J. 2019.
\newblock MIWAE: Deep generative modelling and imputation of incomplete data
  sets.
\newblock In \emph{International conference on machine learning}, 4413--4423.
  PMLR.

\bibitem[{Moreno, Ho, and Vasconcelos(2003)}]{moreno2003kullback}
Moreno, P.; Ho, P.; and Vasconcelos, N. 2003.
\newblock A Kullback-Leibler divergence based kernel for SVM classification in
  multimedia applications.
\newblock \emph{Advances in neural information processing systems}, 16.

\bibitem[{Muzellec et~al.(2020)Muzellec, Josse, Boyer, and
  Cuturi}]{muzellec2020missing}
Muzellec, B.; Josse, J.; Boyer, C.; and Cuturi, M. 2020.
\newblock Missing data imputation using optimal transport.
\newblock In \emph{International Conference on Machine Learning}, 7130--7140.
  PMLR.

\bibitem[{Nazabal et~al.(2020)Nazabal, Olmos, Ghahramani, and
  Valera}]{nazabal2020handling}
Nazabal, A.; Olmos, P.~M.; Ghahramani, Z.; and Valera, I. 2020.
\newblock Handling incomplete heterogeneous data using vaes.
\newblock \emph{Pattern Recognition}, 107: 107501.

\bibitem[{Nguyen, Wainwright, and Jordan(2010)}]{nguyen2010estimating}
Nguyen, X.; Wainwright, M.~J.; and Jordan, M.~I. 2010.
\newblock Estimating divergence functionals and the likelihood ratio by convex
  risk minimization.
\newblock \emph{IEEE Transactions on Information Theory}, 56(11): 5847--5861.

\bibitem[{Nowozin, Cseke, and Tomioka(2016)}]{nowozin2016f}
Nowozin, S.; Cseke, B.; and Tomioka, R. 2016.
\newblock f-gan: Training generative neural samplers using variational
  divergence minimization.
\newblock \emph{Advances in neural information processing systems}, 29.

\bibitem[{P{\'e}rez-Cruz(2008)}]{perez2008kullback}
P{\'e}rez-Cruz, F. 2008.
\newblock Kullback-Leibler divergence estimation of continuous distributions.
\newblock In \emph{2008 IEEE international symposium on information theory},
  1666--1670. IEEE.

\bibitem[{Schennach(2020)}]{schennach2020mismeasured}
Schennach, S.~M. 2020.
\newblock Mismeasured and unobserved variables.
\newblock In \emph{Handbook of Econometrics}, volume~7, 487--565. Elsevier.

\bibitem[{Wang, Kulkarni, and Verd{\'u}(2005)}]{wang2005divergence}
Wang, Q.; Kulkarni, S.~R.; and Verd{\'u}, S. 2005.
\newblock Divergence estimation of continuous distributions based on
  data-dependent partitions.
\newblock \emph{IEEE Transactions on Information Theory}, 51(9): 3064--3074.

\bibitem[{Wang, Kulkarni, and Verd{\'u}(2006)}]{wang2006nearest}
Wang, Q.; Kulkarni, S.~R.; and Verd{\'u}, S. 2006.
\newblock A nearest-neighbor approach to estimating divergence between
  continuous random vectors.
\newblock In \emph{2006 IEEE International Symposium on Information Theory},
  242--246. IEEE.

\bibitem[{Yoon, Jordon, and Schaar(2018)}]{yoon2018gain}
Yoon, J.; Jordon, J.; and Schaar, M. 2018.
\newblock Gain: Missing data imputation using generative adversarial nets.
\newblock In \emph{International conference on machine learning}, 5689--5698.
  PMLR.

\bibitem[{Yoon and Sull(2020)}]{yoon2020gamin}
Yoon, S.; and Sull, S. 2020.
\newblock GAMIN: Generative adversarial multiple imputation network for highly
  missing data.
\newblock In \emph{Proceedings of the IEEE/CVF conference on computer vision
  and pattern recognition}, 8456--8464.

\end{thebibliography}

\end{document}